УДК 004.853, 004.55

*О.В. Палагін, К.С. Малахов, В.Ю. Величко, О.С. Щуров*

# ПРОЕКТУВАННЯ ТА ПРОГРАМНА РЕАЛІЗАЦІЯ ПІДСИСТЕМИ СТВОРЕННЯ ТА ВИКОРИСТАННЯ ОНТОЛОГІЧНОЇ БАЗИ ЗНАНЬ ПУБЛІКАЦІЙ НАУКОВОГО ДОСЛІДНИКА

Створення засобів підтримки наукових досліджень завжди було і є одним із центральних напрямків розвитку інформатики. Головними особливостями сучасного моменту розвитку наукових досліджень є трансдисциплінарний підхід і глибока інтелектуалізація всіх етапів життєвого циклу постановки і вирішення наукових проблем. В роботі розглядаються теоретичні і практичні аспекти розробки перспективних комплексних знання-орієнтованих інформаційних систем та їх компонентів, проведено аналіз існуючих наукових інформаційних систем та синтез загальних принципів побудови "Інструментального комплексу робочого місця наукового дослідника" та його компонентів. Спроектовані та розроблені функціональні компоненти знання-орієнтованої інформаційної системи "Інструментальний комплекс робочого місця наукового дослідника", зокрема функціональні моделі і програмна реалізація підсистеми створення та використання онтологічної бази знань наукового дослідника, як компонент персоніфікованої бази знань наукового дослідника. Дослідження в сучасних умовах дії парадигми e-Science вимагає об'єднання ресурсів наукової спільноти й інтенсивного обміну результатами досліджень, що може досягатись за рахунок використання наукових інформаційних систем. "Інструментальний комплекс робочого місця наукового дослідника" дозволяє вирішувати завдання конструктивізації та формалізації представлення знань, що отримуються в процесі досліджень, та колективної взаємодії співвиконавців.
Ключові слова: семантичний пошук, онтологія, Semantic Web, персоніфікація пошуку.

## Вступ

На поточний стан існує велика кількість наукових інформаційних систем (НІС), головне завдання яких – забезпечити науковою інформацією дослідників і наукові колективи. Наприклад, в університетах багатьох розвинутих країн світу існують інформаційні системи, в яких зберігаються дані про наукові проекти та про виконавців проектів. Однак такі інформаційні системи не задовольняють вимогам широкого доступу до інформації залежно від ролі споживача цієї інформації (вчений, викладач, керівник, інвестор та ін.). Системи бібліотечних установ, які здійснюють реферування публікацій, дозволяють зберігати інформацію про статті, про персони, дозволяють створювати тематичні каталоги. Але жодна з наявних організацій не має достатніх людських ресурсів для збору й аналізу інформації з усіх журналів і Інтернет ресурсів. Тому необхідна комплексна автоматизація накопичення інформації з багатьох джерел. Вирішення такого завдання потребує інтеграції даних з різноманітних інформаційних систем про наукові дослідження та публікації.

В рамках програми інформатизації НАН України за проектом "Розробка методологічних та технологічних засад побудови архітектури знання-орієнтованої інформаційної системи підтримки діяльності наукового дослідника" [1] в Інституті кібернетики імені В.М. Глушкова НАН України розроблені технічні вимоги та архітектура комплексної інформаційної системи наукових досліджень (НД), та запроектовані її компоненти для практичної реалізації у вигляді "Інструментального комплексу робочого місця наукового дослідника" (ІКРМ НДк). На основі отриманих результатів розпочато наступний проект в рамках програми інформатизації НАН України "Створення проблемно-орієнтованих систем онтологічного аналізу і синтезу складних об'єктів нової техніки", основними задачами якого є практична розробка та апробація функціональних компонентів знання-орієнтованого ІКРМ НДк, зокрема, підсистеми створення та використання онтологічної бази знань публікацій наукового дослідника (ОБЗП НДк).

# Аналіз існуючих наукових інформаційних систем

В англомовному середовищі для позначення наукових інформаційних систем з'явився стійкий термін – CRIS (Current Research Information System) [2], що визначає інформаційну систему для доступу до наукової й академічної інформації. Важливо підкреслити, що у визначенні CRIS також вказується, що CRIS призначені не тільки для безпосереднього доступу до інформаційних ресурсів, але й для [3]:

• спрощення доступу до національних служб наукової й технічної інформації;

• ідентифікації головних джерел наукової інформації й оцінки можливостей доступу до цих джерел;

• розвитку мережних інструментів, що дозволяють здійснювати доступ до джерел наукової інформації.

Створено EuroCRIS [4] – професійну організацію дослідників, керівників наукових груп і інститутів країн Європейського Союзу по наукових інформаційних системах. Головне завдання EuroCRIS – поширення інформації в області наукових інформаційних систем; створення тезаурусів, стандартів і методології створення НІС; поширення технологій, необхідних для створення НІС.

Основні категорії користувачів сучасних НІС та їх інформаційні потреби наведено в [5]. Ці категорії користувачів та їх потреби визначають основні види інформаційних ресурсів, з якими працюють НІС: звіти про виконану роботу, результати проектів, персональна інформація, публікації, організації, проекти, наукові результати, технології, патенти, програми фондів, експертні оцінки, електронні бібліотеки, енциклопедії та тлумачні словники, веб-сайти, списки розсилки, соціальні мережі, бази даних, обчислювальні ресурси, нормативні документи, освітні й музейні ресурси. Також категорії користувачів і їх потреби визначають основні види сервісів, які надають НІС: повторне використання наукових розробок, методологій, технологій; пошук та цільове поширення інформації; встановлення горизонтальних зав'язків між організаціями та колективами науковців; архівне зберігання інформації; підтримка навчального процесу; формування запитів як дослідників, так і користувачів наукових знань; забезпечення роботи аналітичних служб.

Аналіз вимог до CRIS з управління наукою проаналізовано в [6], де описано типи діяльності, інформаційні потреби в управлінні наукою й види ресурсів, з якими має працювати CRIS для керівників, аналітиків та фінансових фондів. В роботі [7] наведено вимоги до CRIS для розповсюдження технологій, організації наукових програм і роботи фондів з фінансування наукових досліджень. Зазначено важливість CRIS систем для організації спільної роботи вчених та підтримки інформаційної роботи фондів з фінансування. Описано базовий життєвий цикл наукових програм і інформаційні потреби учасників на кожній фазі циклу.

У порталі CORDIS [8] виділено основні види використання наукових порталів для дослідників:

• доступ до актуальної інформації про досягнення в науці;

• ідентифікація організацій, що фінансують проекти для дослідження й створення технологій;

• пошук партнерів для організації наукової діяльності;

• формування наукових колективів, що можуть складатись з окремих незалежних груп;

• експорт технологій і їх результатів.

В області наукової комунікації в мережі Інтернет сьогодні використовуються: сервіси вербальної безпосередньої взаємодії як для спілкування за допомогою персонального комп'ютера, так і мобільних пристроїв зв'язку; платформи для проведення вебінарів; сервіси опосередкованого спілкування – електронна пошта; корпоративні месенджери; соціальні мережі. Такі сервіси за наявності швидкісних каналів зв'язку дозволяють здійснювати оперативну наукову комунікацію.

До НІС можна віднести системи, у яких поєднуються технології як спеціалізованої соціальної мережі, так і звичайного файлового сховища [9]. Комерційні проекти подібних систем орієнтовані на використання однієї з наступних бізнес-моделей:

• продавати дані своїх відвідувачів або передплатників рекламодавцям, що забезпечує цільову рекламу, яка зазвичай коштує дорожче звичайної (ResearchGate) [10];

• проводити аналітику по завантаженому контенту й підбирати необхідні дослідження за окрему плату (Academia.edu) [11];

• надавати додаткові платні сервіси для зберігання матеріалів, організовувати дискусійні площадки (Mendeley) [12].

Відповідно до потреб саме представників наукового середовища, можна виділити наступні основні можливості Academia.edu і ResearchGate, які надаються зареєстрованим користувачам: завантажувати в систему свої наукові тексти, які будуть доступні для інших користувачів; користуватись аналітичними можливостями систем, зокрема, мати доступ до статистики перегляду свого профілю, публікацій і завантаження кожної публікації; виконувати пошук за науковими інтересами користувачів і встановлювати з ними академічні контакти. Сервіси підтримують персоніфіковане формування стрічки новин, в якій повідомлення про нові публікації будуть з'являтись від користувачів, з якими встановлено контакти в самому сервісі або при співпадінні ключових слів статей і інтересів користувачів, що сприяє цільовому поширенню результатів наукових досліджень. Статистика перегляду публікацій дозволяє визначити країну походження запиту на перегляд публікації та інші параметри запиту.

Однак, сервіс Academia.edu не розрахований на комплексну інформаційну підтримку проведення наукових досліджень, а більшість користувачів використовують сервіс для розміщення в системі своїх опублікованих статей. Сервіс ReserchGate має більше різноманітних засобів підтримки діяльності дослідника.

Наприклад, за тематикою наукових інтересів користувачу може бути надана можливість виступити як науковому експерту: відповісти на питання, які задані іншим користувачем. Також доступна роль рецензента публікацій, розміщених у системі (пошук здійснюється за назвами статей). У системі надається можливість пошуку цитування своїх статей й відображення інформації про цитування. Особливістю сервісу ReserchGate є можливість організації проектів із залученням до спільної роботи над ними інших користувачів системи, з якими встановлено академічні контакти. Ця можливість реалізована через створення простору в якому можна розміщувати файли, проводити обговорення з іншими учасниками проекту у вигляді коментарів.

Інтернет-сервіси (Academia.edu, ReserchGate) багато в чому націлені на формування так званого "імені" вченого і надання послуг з видачі вакансій за науковими інтересами користувачів, а також з перегляду профілів користувачів потенційними роботодавцями, фондами, керівниками науково-дослідних колективів.

## Принципи побудови "Інструментального комплексу робочого місця наукового дослідника"

Науковими співтовариствами створюються мережі, що поєднують цифрові бібліотеки, файлові сховища, веб сервери з науково значимою інформацією. Головним об'єктивним фактором, який необхідно враховувати при розробці і використанні робочого місця українського наукового дослідника, є обмеженість фінансових ресурсів, що виділяються на розробку та експлуатацію програмного забезпечення робочого місця, що приводить до наступних наслідків: неможливість залучення до розробки необхідної кількості спеціалістів; мінімізація витрат на освоєння, експлуатацію та оновлення програмного забезпечення; розробка системи на протязі декількох років колективом виконавців, склад якого змінюється; необхідність максимального повторного використання програмного забезпечення, що вже розроблено. Враховуючи наведене вище, визначимо загальні принципи, на яких має будуватись інформаційна система ІКРМ НДк та її компоненти.

• *Модульність.* Розробка будь-якої інформаційної системи, тим більше такої, що націлена на рішення комплексу різних завдань, має базуватись на модульному принципі [9]. Реалізація кожного окремого завдання вирішується в рамках створення окремого програмного модуля, який у будь-який момент можна підключити до системи, що дозволяє нарощувати її функціональні можливості. При модульному принципі побудови НІС немає необхідності вводити систему в експлуатацію після остаточної розробки всіх її сервісів і компонентів – вона може використовуватись спочатку з мінімальним набором функцій та поступово розвиватись, додаючи необхідну функціональність. Модульна побудова системи дозволяє оперативно виконувати завдання налагодження, тестування та заміни як окремих модулів, так і всієї системи в цілому.

• *Сервіс-орієнтована архітектура.* Вона може розглядатись як стиль архітектури інформаційних систем, який дозволяє створювати системи, що побудовані шляхом комбінації окремих незалежних програм [13]. Програми, такі як веб-сервіси, можуть викликатись іншими програмами, які виступають як клієнти або споживачі цих сервісів. Розробнику не потрібно знати, як працює програма, необхідно лише дотримуватись угоди про інтерфейс для звернення до сервісу, та форматів вхідних і вихідних даних. Інтерфейс сервісу повинен не залежати від платформи. Сервіс-орієнтована архітектура реалізує масштабованість сервісів, забезпечує скорочення часу реалізації проекту, підвищення продуктивності його розробки та впровадження.

• *Кросплатформеність.* Дослідник повинен мати доступ до сервісів, які забезпечує НІС, з різних типів пристроїв, що працюють під керуванням різних операційних систем. Одним з засобів досягнення цієї мети може бути доступ до сервісів через адаптивний веб-інтерфейс.

• *Розробка програмного забезпечення за шаблоном Модель-вигляд-контролер (Model-view-controller MVC)* – архітектурний шаблон, що використовується під час проектування та розробки програмного забезпечення і поділяє програму систему на три частини: модель даних, вигляд даних та засоби обробки даних. Шаблон MVC призначений для відокремлення даних (модель) від інтерфейсу користувача (вигляду) таким чином, щоб зміни інтерфейсу користувача мінімально впливали на роботу з даними, а зміни в моделі даних могли здійснюватися без змін інтерфейсу користувача. Перевага шаблону – гнучкий дизайн програмного забезпечення, який полегшує подальші зміни чи розширення програм, надає можливість повторного використання окремих компонентів. Крім того, використання цього шаблону у великих системах призводить до певної впорядкованості їхньої структури, робить їх більш зрозумілими, завдяки зменшенню складності окремих компонентів.

• *Взаємодія між компонентами на основі JSON моделі даних та архітектурного стилю взаємодії компонентів розподіленого програмного забезпечення REST* (скор. англ. Representational State Transfer, "передача репрезентативного стану"). JSON – стандарт побудови мов розмітки ієрархічно структурованих даних для обміну між різними додатками, зокрема, через Інтернет. Стандарт визначає метамову, на основі якої, шляхом запровадження обмежень на структуру та зміст документів, визначаються специфічні, предметно-орієнтовані мови розмітки даних. Автор документа створює його структуру, будує необхідні зв'язки між елементами, й використовує ті команди, які задовольняють його вимогам, і домагається такого типу розмітки, який потрібен йому для виконання операцій з документами. JSON дозволяє також здійснювати контроль за коректністю даних, що зберігаються в документах, робити перевірки ієрархічних співвідношень усередині документа і встановлювати єдиний стандарт на структуру документів, що вміщують будь-які дані. Створи-вши правильну структуру механізму обміну інформації на початку роботи над проектом системи, можна уникнути в майбутньому багатьох проблем, пов'язаних з несумісністю форматів даних, які використовуються різноманітними компонентами системи.

● *Максимальне використання вільного та безкоштовного програмного забезпечення у комплексі НІС.* Особливістю вільного програмного забезпечення є відкритість кодів програм, відсутність витрат користувачів на придбання ліцензій, можливість вільного копіювання та розповсюдження програм, безкоштовність (або невисока вартість розповсюдження копії при використанні програми у комплексному рішенні), можливість модифікації програм і значне скорочення витрат на розробку рішень, необхідних для розв'язання конкретних задач використання НІС. Вільне програмне забезпечення, або його модифіковані версії, можна поширювати як безкоштовно, так і на комерційній основі. Використання вільного програмного забезпечення дозволяє суттєво заощаджувати кошти на етапах розробки та впровадження НІС.

● *Використання хмарних обчислень.* Хмарні обчислення (cloud computing) можуть забезпечити взаємодію між різними НІС через Інтернет, при оптимальному розподілі навантаження між локальними й віддаленими серверами. У хмарних обчисленнях комп'ютерні ресурси й потужності надаються користувачеві як інтернет-сервіси для обробки даних. Користувач має доступ до власних даних, але не повинен піклуватися про інфраструктуру, операційну систему й властивості програмного забезпеченні, з яким працює.

● *Зберігання даних на основі абстракції сховища даних.* Абстрагування програмних модулів від реалізації технології зберігання даних дозволяє адміністратору системи самому обирати тип сховища даних, яке максимально пристосоване для цілей роботи конкретної конфігурації НІС.

# Персоніфікована база знань наукового дослідника

Персоніфікована база знань наукового дослідника (ПфБЗ НДк) – це компонент "Інструментального комплексу робочого місця наукового дослідника" – відкрита комп'ютерна база знань, побудована автоматизованим способом за допомогою заданого, спеціалізованого інструментарію шляхом комплексної обробки авторських науково-технічних матеріалів, що опубліковані у відкритому доступі, чи отриманні від НД у вигляді експертних знань.

ПфБЗ НДк представлена у комп'ютерній формі у вигляді онтологічної бази знань і формально описана на деякій мові першого порядку. ПфБЗ НДк призначена для [1]:

● опису максимального наближення до точного онтографічного подання знань НД;

● структурно-семантичного аналізу НДк своїх знань, їх поповнення, оновлення та підтримки в актуальному стані, в том числі для підготовки до наступних НД;

● інформаційного забезпечення й орієнтації: розвитку персоніфікованих знань для побудови морфологічних таблиць онтографічної технології наукової та технічної творчості; для інструменту автоматизації написання наукових статей, заявок на отримання патентів та інших науково-технічних документів;

● онтолого-орієнтованого інформаційно-довідкового забезпечення НДк;

● уточнення, розширення та розвитку загального фрагмента ПфБЗ НДк і онтології основної предметної дисципліни, в якій проводяться НД;

● формування тематичної спрямованості нових НД;

● планування розробки нових наукових теорій для вирішення прикладних задач;

● формулювання запитів на отримання нової, тематично спрямованої інформації;

● підтримки виконання конкретного наукового проекту.

Архітектура ПфБЗ НДк представлена сукупністю блоків та підсистем: онтологічна база знань предметної області (ПрО), підсистема створення та використання онтологічної бази знань публікацій наукового дослідника (ОБЗП НДк), блок системного аналізу, об'єкти інтелектуальної власності автора (зокрема, патенти), лінгвістичний корпус ПрО, інформація про поточний проект.

## Проектування функціональної моделі підсистеми онтологічної бази знань публікацій наукового дослідника

Основною підсистемою ПфБЗ НДк є онтологічна база знань публікацій наукового дослідника, що включає у себе додану НДк, зібрану в мережі Інтернет спеціалізованим "роботом-павуком" та отриману від інших CRIS-систем інформацію з різних предметних областей, та кластери онтологій та онтологічних структур, що в напівавтоматичному режимі побудовані програмним забезпеченням на основі отриманої інформації.

Для проектування функціональної моделі підсистеми ОБЗП НДк була використана загальноприйнята стандартна методологія і мова функціонального моделювання UML. UML – де-факто стандарт об'єктно-орієнтованої візуальної мови моделювання. В UML є три основні види моделей [14]: статична модель (static model), динамічна модель (dynamic model), фізична модель (physical model).

Статична модель описує елементи системи та їх відношення (класи, атрибути, оператори). Однією з реалізацій статичної моделі є діаграма класів (class diagram). Динамічна модель описує поведінку системи, наприклад, зміна програмних сутностей (software entities) під час виконання додатка. До динамічних моделей відносяться: діаграма прецедентів, варіантів використання (use case diagram); діаграма активності (activity diagram). Фізична модель відображає незмінну структуру програмних сутностей, зокрема, файлів програмного коду, бібліотек, що виконуються та відношення між ними. До підсистеми ОБЗП НДк входять наступні сутності, блоки та функції:

- сутність "Науковий дослідник";
- блок "Підсистема управління колективним проектом";
- блок керування графічним редактором онтологічних структур;
- блок керування пошуком наукової інформації в зовнішніх джерелах (в мережі Internet);
- блок обробки лінгвістичної інформації для онтологічної репрезентації текстів – ІТ-платформа ТОДОС (Трансдисциплінарні Онтологічні Діалоги Об'єктно-орієнтовних Систем) [15];
- блок документно-орієнтованого сховища даних – сервер системи керування базами даних (СКБД) MongoDB;
- функція графічного представлення онтологічних структур, онтологій та наукової інформації;
- функція запису/зчитування інформації у вигляді оригіналів публікацій НДк, JSON та XML документів, онтологічних структур, онтологій в СКБД MongoDB;
- функція локальної роботи з даними;
- функція повнотекстового пошуку по публікаціям НДк, онтологіям та онтологічним структурам;
- функція індексації та розмітки оригіналів публікацій НДк для повнотекстового пошуку;
- функція семантичного аналізу текстових документів;
- функція автентифікації/авторизації НДк;
- функція автоматичного машинного перекладу публікацій НДк з англійської та російської на українську мову;
- функція лематизації словоформ для української мови;
- функція екстракту тексту з документів форматів pdf, doc та docx.

Функціональні моделі ІКРМ НДк та його компонентів, а саме динамічні моделі у вигляді UML-діаграм варіантів використання, наведені в [1]. Отримані результати проектування функціональних моделей компонентів ІКРМ НДк дозволили перейти до їх програмної реалізації, зокрема підсистеми ОБЗП НДк.

## Програмна реалізація підсистеми онтологічної бази знань публікацій наукового дослідника

Програмна реалізація підсистеми ОБЗП НДк виконана з дотриманням наведених вище принципів розробки комплексних інформаційних систем. Для практичної програмної реалізації ОБЗП НДк обрано формат веб-сервісів та веб-застосунку, що дозволило створити гнучке та багатофункціональне програмне забезпечення з сервіс-орієнтованою розподіленою архітектурою та задіянням "хмарних" технологій. За допомогою цих сервісів та веб-застосунку можна ефективно будувати кластери онтологій, здійснювати пошук, зберігання, синтаксично-семантичний аналіз текстових документів та інше.

Основою розробки програмної реалізації став стек технологій серверного програмного забезпечення MEAN (абр. від MongoDB, Express.js, Angular.js, Node.js). Стек технологій MEAN відображає сучасний підхід до веб-розробки: на кожному рівні архітектури застосунку, від клієнта до сервера і персистентності, застосовується одна й та сама мова програмування, зокрема в даному випадку – JavaScript. В існуючій експериментальній версії програмної реалізації замість JavaScript-фреймворка Angular.js були використані JavaScript-бібліотека jQuery та підхід до побудови користувацьких інтерфейсів веб-застосунків – AJAX (Asynchronous JavaScript And XML). На рисунку показано узагальнену структурну схему програмної реалізації ОБЗП НДк з використанням стека MEAN та зміною Angular.js на jQuery та AJAX. Детальний опис стека MEAN та його компонентів представлено у [16, 17]. Розроблене програмне забезпечення умовно можна поділити на дві складові частини:

1. Веб-сервіс у вигляді RESTful API, що реалізує функціональне наповнення ОБЗП НДк та працює, як самостійний сервіс, так і інтегрований з веб-застосунком ОБЗП НДк.

2. Веб-застосунок, який забезпечує взаємодію НДк з системою та використання її сервісів.

Клієнтська частина реалізована у вигляді односторінкового веб-застосунку (англ. Single-page application, SPA), що забезпечує НДк досвід, близький до користування настільною програмою. Інтерфейс користувача веб-застосунку складається з вікна виводу результату та набору логічно поділених вкладок: пошук з ОБЗП НДк, завантаження нових даних, налаштувань сервісу з формою, яка за рахунок авторизації користувача та блоку користувача дозволяє користувачеві, виходячи з власних потреб, індивідуально налаштовувати інтерфейс та відображати лише необхідні для роботи ОБЗП НДк, що розташовуються з окремих вкладках форми. Такий формат відображення сервісу при потребі дозволяє додавати нові елементи сервісу та розширяти його функціональність у майбутньому, не привносячи значних змін у HTML та CSS код сторінки та зовнішній вид веб-сервісу, що спрощує його розвиток та обслуговування у майбутньому та виключає адаптаційні складнощі постійних користувачів веб-сервісу при переході на його оновлену версію.

Сервісна частина ОБЗП НДк реалізована з використанням СКБД MongoDB, що здійснює керування наборами документів у форматі JSON. Для реалізації керування ОБЗП НДк в СКБД MongoDB була розроблена JSON специфікація опису метаданих публікацій автора. Логіка роботи сервісу, обробка запитів користувача, пошук та аналіз даних на веб-сервері побудовано на програмній платформі з відкритим початковим кодом Node.js, яка працює на рушії JavaScript від Google – V8 та призначена для відокремленого виконання високопродуктивних мережних застосунків на мові JavaScript. Node.js характеризується асинхронною однопотоковою моделлю виконання запитів, засновану на обробці подій у неблокуючому режимі.

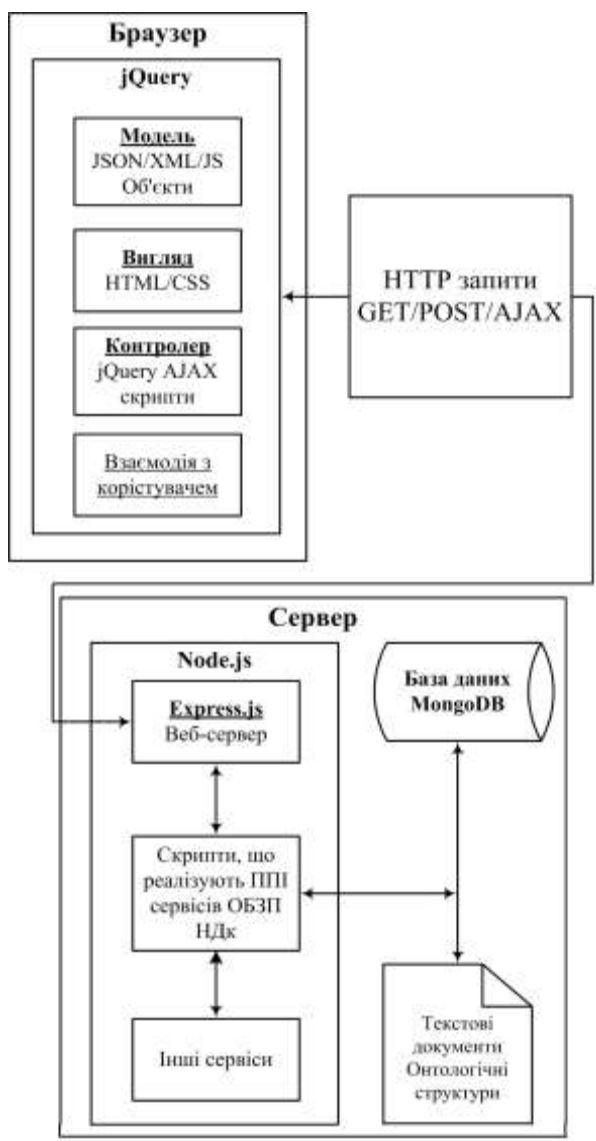

Рисунок. Узагальнена структурна схема програмної реалізації ОБЗП НДк з використанням стека MEAN

ОБЗП НДк працює у тестовому режимі та доступна для користування за посиланням: http://icybcluster.org.ua:33145/ .

## Висновки

Створення засобів підтримки наукових досліджень завжди було і є одним із центральних напрямків розвитку практичної інформатики. Головними особливостями сучасного моменту розвитку наукових досліджень є трансдисциплінарний підхід і глибока інтелектуалізація всіх етапів життєвого циклу постановки і вирішення наукових проблем. Дослідження в сучасних умовах дії парадигми e-Science з дуже складними та масштабними експериментами й спостереження-ми, вивченням явищ на перетині галузей науки (міждисциплінарні та трансдисциплінарні дослідження), вимагає об'єднання ресурсів всієї наукової спільноти, розподілення всього обсягу досліджень між окремими науковцями та науковими колективами й інтенсивного обміну результатами досліджень. Для реалізації цих можливостей спроектовані та розроблені функціональні компоненти знання-орієнтованої інформаційної системи ІКРМ НДк, зокрема, функціональні моделі і програмна реалізація підсистеми створення та використання ОБЗП НДк, як компоненти ПфБЗ НДк.

**Про авторів:**

*Палагін Олександр Васильович,*
доктор технічних наук, професор,
академік НАН України.
Кількість наукових публікацій в
українських виданнях – 290.
Кількість наукових публікацій в
зарубіжних виданнях – 45.
H-index: Google Scholar –15;
Scopus – 3.
http://orcid.org/0000-0003-3223-1391,

*Малахов Кирило Сергійович,*
молодший науковий співробітник.
Кількість наукових публікацій в
українських виданнях – 27.
Кількість наукових публікацій в
зарубіжних виданнях – 2.
H-index: Google Scholar – 2.
http://orcid.org/0000-0003-3223-9844,

*Величко Віталій Юрійович,*
кандидат технічних наук, доцент,
старший науковий співробітник.
Кількість наукових публікацій в
українських виданнях – 73.
Кількість наукових публікацій в
зарубіжних виданнях – 25.
H-index: Google Scholar – 7; Scopus – 1.
http://orcid.org/0000-0002-7155-9202,

*Щуров Олександр Сергійович,*
інженер програміст 1 категорії.
Кількість наукових публікацій в
українських виданнях – 4.
H-index: Google Scholar – 1.
http://orcid.org/0000-0002-0449-1295

**Місце роботи авторів:**

Інститут кібернетики
імені В.М. Глушкова НАН України.
03680, Київ-187,
проспект Академіка Глушкова, 40.
Тел.: (044) 526 3348.

# Література